\documentclass[conference]{IEEEtran}
\IEEEoverridecommandlockouts
\usepackage{etoolbox}
\makeatletter
\patchcmd{\@makecaption}{\scshape}{}{}{}
\makeatother

\usepackage{siunitx}
\usepackage{mathtools}
\usepackage{dsfont}
\usepackage{tipa}
\usepackage{booktabs}
\usepackage{multirow}
\usepackage{gensymb}
\usepackage{cite}
\usepackage{amsmath,amssymb,amsfonts}
\usepackage{algorithmic}
\usepackage{graphicx}
\usepackage{textcomp}
\usepackage{wasysym}
\usepackage{pifont}
\usepackage{xcolor,colortbl}
\usepackage{hyperref}
\usepackage{makecell}
\definecolor{inchworm}{rgb}{0.7, 0.93, 0.36}
\definecolor{gold}{rgb}{1.0, 0.84, 0.0}
\definecolor{lemon}{rgb}{1.0, 0.97, 0.0}
\definecolor{mypurple}{RGB}{94, 47, 152}
\definecolor{first}{RGB}{119, 160, 249}
\definecolor{second}{RGB}{209, 224, 255}
\usepackage{contour}

\contourlength{0.1pt} 
\usepackage{comment}
\newcommand{\fs}{ \cellcolor{first}\bf }   
\newcommand{\nd}{ \cellcolor{second}\underline }      

\newcommand{\cmark}{\textcolor{green}{\ding{51}}} 
\newcommand{\xmark}{\textcolor{red}{\ding{55}}} 
\def\BibTeX{{\rm B\kern-.05em{\sc i\kern-.025em b}\kern-.08em
    T\kern-.1667em\lower.7ex\hbox{E}\kern-.125emX}}
\begin{document}
\pagestyle{plain} 
\title{
Lost \& Found: Tracking Changes from Egocentric Observations in 3D Dynamic Scene Graphs}

\author{\IEEEauthorblockN{Tjark Behrens$^{1}$, René Zurbrügg$^{1}$, Marc Pollefeys$^{1,2}$, Zuria Bauer$^{1,\dagger}$, Hermann Blum$^{1,3,\dagger}$}

\thanks{$^1$ETH Zürich, $^2$Microsoft, $^3$Uni Bonn, $\dagger$ denotes equal advising.}
\thanks{Corresponding Author: Tjark Behrens $<$\href{mailto:tbehrens@ethz.ch}{tbehrens@ethz.ch}$>$} \thanks{This research is partially supported by the ETH AI Center, ETH Foundation Project 2025-FS-352, the SNSF Advanced Grant 216260 and the Lamarr Institute for Machine Learning and Artificial Intelligence. We thank Meta Project Aria for providing their hardware.}
}

\maketitle

\begin{abstract}
Recent approaches have successfully focused on the segmentation of static reconstructions, thereby equipping downstream applications with semantic 3D understanding. However, the world in which we live is dynamic, characterized by numerous interactions between the environment and humans or robotic agents. Static semantic maps are unable to capture this information, and the naive solution of rescanning the environment after every change is both costly and ineffective in tracking e.g. objects being stored away in drawers. With Lost \& Found we present an approach that addresses this limitation. Based solely on egocentric recordings with corresponding hand position and camera pose estimates, we are able to track the 6DoF poses of the moving object within the detected interaction interval. These changes are applied online to a transformable scene graph that captures object-level relations. Compared to state-of-the-art object pose trackers, our approach is more reliable in handling the challenging egocentric viewpoint and the lack of depth information. It outperforms the second-best approach by 34\% and 56\% for translational and orientational error, respectively, and produces visibly smoother 6DoF object trajectories. In addition, we illustrate how the acquired interaction information in the dynamic scene graph can be employed in the context of robotic applications that would otherwise be unfeasible: We show how our method allows to command a mobile manipulator through teach \& repeat, and how information about prior interaction allows a mobile manipulator to retrieve an object hidden in a drawer. Code, videos and corresponding data are accessible at \href{https://behretj.github.io/LostAndFound/}{behretj.github.io/LostAndFound/}.
\end{abstract}



\section{Introduction}

Dynamic scene understanding is a fundamental requirement for autonomous robots operating in diverse environments, such as homes, warehouses, or farms. While substantial progress has been made in semantic scene understanding, most research to date has focused on static scenes.
However, assuming environments to be static severely limits real-world applicability as everyday environments such as living rooms are inherently dynamic since objects are moved and furniture might be rearranged.

Previous works in reconstruction of dynamic scenes usually investigate the joint problem of reconstructing the 3D geometry of all objects and their movements together. To achieve this, they require a prohibitive number of calibrated cameras~\cite{luiten2024dynamic}, ignore observations during object movement~\cite{khronos,dynablox} or are computationally limited to small tabletop scenes~\cite{midfusion}.
To address these limitations, we propose to decompose the problem of dynamic semantic mapping into two tasks: (i) Estimate the geometry of each object and (ii) accurately track its pose over time. We leverage a prior, \emph{static} reconstruction of the environment for the first task, for which many methods already exist. For the second task, our method enables the tracking and mirroring of user interactions in this environment. Our environment reconstruction, represented as a dynamic scene graph, provides a compact and flexible framework for capturing the ongoing dynamics within a scene that is later on also accessible by robots.

Our approach demonstrates the ability to perform tasks that are typically beyond the capabilities of static semantic maps, such as retrieving objects from a drawer after a human has placed them there. Static maps are inherently limited in such scenarios, as they cannot capture the contents of a closed drawer or other occluded environments.
\begin{figure}[t]
    \centering
    \includegraphics[width=0.95\linewidth]{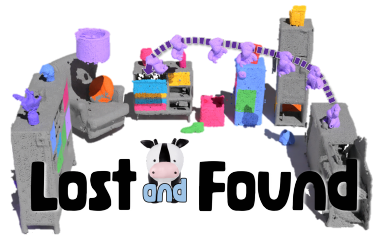}
    \caption{\textbf{Object tracking}: Our method allows to track the \emph{dynamic} information associated with the pick-and-place action of the cow toy (\textbf{\textcolor{mypurple}{purple}}). The figure presents a series of snapshots of the employed scene graph data structure at varying time points. The object trajectory extends from the right shelf to the top-left drawer of the cabinet, as indicated by the dotted arrow.} 
    \label{fig:teaser_fig}
    \vspace*{-8mm}
\end{figure}
To achieve this, we capture user interactions in dynamic environments through egocentric cameras, specifically utilizing Aria glasses~\cite{projectaria}. By registering the Aria glasses to the scene reconstruction, we accurately track their pose and, through the glasses' cameras, the 3D positions of both hands, as long as they remain within the field of view. Also using the egocentric camera, we predict interactions between hands and objects, which can then be matched with the 3D poses of object instances in the scene. This integrated setup allows us to robustly track pick-and-place interactions with objects as well as furniture interactions, such as the opening and closing of drawers, performed by the user wearing the glasses.

The resulting dynamic scene graph representation is directly accessible to robots. Through its high level of abstraction, we effectively decouple the environment from both visual observations and human embodiment.
We illustrate this by replaying a pick-and-place action of a human with a robot in a teach-and-repeat fashion, as well as retrieving an object with a robot from a closed drawer that was previously placed there by a human. In summary, our contributions are: 
\begin{itemize}
    \item We introduce a method to track objects in egocentric video by fusing hand positions, spatial object tracks, and a prior static reconstruction.
    \item We spatiotemporal reconstruct object and furniture interactions in a robot-accessible, dynamic 3D scene graph.
    \item We thoroughly evaluate and demonstrate our method in challenging real-world scenarios.
\end{itemize}

\section{Related Work}

\noindent{\textbf{3D Instance Segmentation:}} Instance segmentation of point clouds identifies and classifies individual objects within a scene. This is essential in dynamic settings like ours, where each of these parts may move independently.

Coarse-to-fine-based solutions infer rough object detections and then repetitively refine the estimate to an accurate segmentation.
Instead of initial 3D bounding box estimates, Spherical Mask~\cite{spherical} proposes 3D polygons based on spherical coordinates as an instance representation. These estimates are treated as a soft reference and not as a perfect initialization.

Transformer-based Architectures have been established as the state-of-the-art in 3D instance segmentation of point clouds. Mask3D~\cite{mask3d} proposes a sparse 3D convolutional U-Net architecture to encode the point cloud features. Iterative refinement steps for the instance queries in the transformer decoder, together with the extracted features, allow for the final prediction of instance masks and semantic labels per point.

Similarly to foundational models for 2D tasks, Unified Frameworks \cite{oneformer3d, unified3d} have been proposed that exploit multi-task learning to learn one large model that performs on par or better than single-task models.



\begin{figure*}[ht!]
    \centering
    \includegraphics[width=1.0\linewidth]{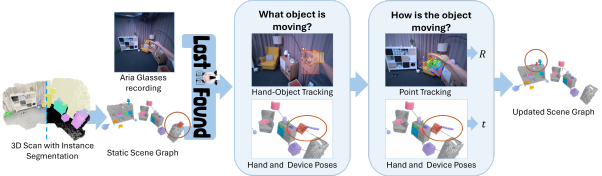}
    \caption{\textbf{Method Overview:} We build a static scene graph that captures object-level relationships, given our initial 3D scan with its semantic instance segmentation. Each Aria glasses recording provides hand positions and device poses. With Lost \& Found, we identify object interactions by locating them in our 3D prior and simultaneously querying a 2D hand-object tracker. At the beginning of such an interaction, we project 3D points of our object instance onto the image plane. A point tracking method keeps track of these 2D feature points in subsequent observations. While the 3D hand location yields an anchor for the object translation, we can apply a robust perspective-n-point algorithm to the known 2D-3D correspondences for each RGB image, to identify the correct 6DoF pose of the object. The scene graph is updated accordingly to reflect the correct state of the current environment. In the example above, the picture frame (red) is carried from the rack on the right to the top of the tall shelf on the left.}
    \label{fig:method-overview}
    \vspace*{-1pt}
\end{figure*}
\enlargethispage{\baselineskip}
\noindent{\textbf{3D Scene Graphs}} essentially split a 3D scene into smaller parts, where each part is a node and the overall structure is kept through properties and edges between nodes. This makes it naturally a more flexible representation than point clouds or voxel grids because moving an object only requires changing a few properties and/or edges. \enlargethispage{\baselineskip}

A common approach is to structure the scene graph in layers. While the root node represents e.g. a complete building, lower levels represent the rooms and objects, respectively. The lowest level may contain camera information of the capturing process or the obtained 3D representation (mesh/point cloud). Top-down connections are established to induce this natural hierarchy. Links within one layer indicate semantic or spatial relations of objects. Multiple methods have been proposed to build such scene graphs from sensory observations~\cite{kimera, s-graphs, hydra}.
In general, these methods have in common that once the scene graph is built, it is not updated anymore, thus, only capturing static information. More recently with the rise of open-vocabulary methods, ConceptGraphs~\cite{conceptgraphs} demonstrate that the semantics of an object allow for informative guessing when an object is not in the location marked in the scene graph. In~\cite{3dvsg}, they extend 3D scene graphs to model the likelihood of long-term semantic scene changes but without capturing any actual interactions.
Dynamic Scene Graphs (DSGs)~\cite{3dsg} is the approach that is the most related to our method as they also distinguish between static and dynamic objects. They restrict DSGs to humans and robots as dynamic objects to observe their trajectories in the scene. Hence, they are capable of answering time-dependent questions in this regard. However, all other objects are modelled as static centroids with bounding boxes. \\
\textbf{Pose Tracking:} Object pose estimation solely based on RGB images remains a difficult problem. Approaches such as Pix2Pose~\cite{pix2pose} predict the 3D location of every image pixel. This yields dense 2D-3D correspondences that allow one to run a RANSAC-based Perspective-n-Point~\cite{pnp} (PnP) to obtain the 6D pose of an object. With recent advances in point tracking \cite{tapir, pips, cotracker}, natural extensions to the 3D domain \cite{spatracker} could benefit this idea.
\enlargethispage{\baselineskip}

When depth is available as an additional signal, the setting becomes less constrained and deep learning-based models excel. The literature distinguishes between model-free and model-based approaches, depending on the availability of a CAD model of the tracked object. In the model-free setting, BundleTrack~\cite{bundletrack} computes an initial coarse pose estimate via robust key point feature matching. This initialization is refined online based on a global pose graph optimization. BundleSDF~\cite{bundlesdf} performs neural 3D reconstruction of the object, to jointly optimize this implicit surface representation and the pose estimations in almost real-time. FoundationPose~\cite{foundationpose} is a recent proposal that reaches the state of the art in both categories, leveraging the advantages of a novel transformer-based architecture and a contrastive learning formulation. 

\noindent{\textbf{Hand-Object Interactions:}} The identification of hand-object interactions represents a powerful tool, given that hands are the primary instrument through which humans interact with their environment. However, the estimation of three-dimensional distances from a single image represents a long-standing challenge, which is pivotal for predicting whether a hand is in contact with an object. Action recognition is further complicated by an egocentric perspective, which may result in the hand obscuring the object of interest, as well as the presence of rapid movements and background clutter. Current works~\cite{Shan20} propose a Faster-RCNN backbone that benefits from the large-scale data set for this purpose. 
\enlargethispage{\baselineskip}

\noindent{\textbf{Dynamic Semantic SLAM:}} Schmid et al.~\cite{khronos} offer a dynamic SLAM formulation that can represent semantic maps that change over time but assume reconstructed objects to remain static over multiple frames while observing them, therefore focusing on changes that occur outside of the camera field-of-view. We in contrast focus only on dynamics that happen during the camera observation. Like us, MID-Fusion~\cite{midfusion} tracks dynamic object movements in front of the camera. While it even performs reconstruction and modelling simultaneously, it also requires RGB-D sensing and is only demonstrated on a tabletop scene.

\section{Method}
\subsection{Problem Formulation}\label{problem}

Given egocentric observations from a head-mounted device with pose, ego-view, and hand position information (for example, from Aria glasses) and the initial state of a scene graph representation, we aim to accurately track and update object locations and relations from human pick-and-place interactions. An overview of our approach is shown in Fig.~\ref{fig:method-overview}.
\vspace{-5pt}
\subsection{Scene Graph Structure}
\enlargethispage{\baselineskip}
We define our scene graph as a set of nodes and edges $G~=~(V, E)$. 
For nodes, we focus this work on the levels of objects and object parts, but our definition is fully compatible with room and building level graph hierarchies as in~\cite{3dsg,hydra,conceptgraphs}. We assign each node $u \in V$ a semantic label and the segmented points belonging to the instance.
While other object-part nodes are possible, we focus on drawers, for which we additionally store a 3D bounding box for the volume of their content.
Without loss of generality, we illustrate the possibility of node relations by adding a 'close to' edge $(u,v) \in E$ for every nearest-neighbor relationship between the centroid of $u$ to $v$.
Additionally, we consider two more specific node relations: (i) 'part of' edges that connect an object part to a parent object, e.g. drawers to a cabinet; (ii) 'contains' edges that connect objects stored inside drawers to the drawer.

The scene graph can be efficiently stored as an indexed list of nodes, an adjacency list for edges, and a k-d-tree containing the 3D node centroids for efficient nearest-neighbor querying.

For updating node relations during an interaction, it is sufficient to check whether the connections of the carried object itself and the ones of its nearest neighbor still remain correct. We update 'close to' edges by running a new nearest neighbor search on the interacted node and all nodes it was connected to with a 'close to' edge. Further, we remove 'contains' edges once the object centroid moves out of the drawer's 3D bounding box or we add 'contains' edges once the centroid moves into the 3D bounding box of the closest drawer node.

\subsection{Scene Graph Initialisation}%
We initialize the set of nodes by running a 3D instance segmentation method~\cite{mask3d} over an initial static point cloud scan of the scene. 
To demonstrate the adaptability of the scene structure, we further incorporate image-based detection into our initial scene graph. Detected objects from RGB images of the 3D scanning trajectory are lifted into the 3D point cloud. This enables us to detect objects that are usually hard to capture in point clouds, such as posters. We integrate the drawer detector of \cite{lemke2024spotcompose} to spawn drawers as individual nodes and add connections to the respective cabinet.
\label{sec:extensions}

\subsection{Tracking of Interactions}
\enlargethispage{\baselineskip}
Given the Aria recording, we process the observations frame by frame in a delayed online manner. We follow a sliding window approach, consisting of an observation buffer $B$ storing past frames and a look-ahead horizon $H$ for future time steps, that imposes a delay of $|H|$ frames on the overall algorithm.
For each frame $k$, we look at an observation $y_k$, which describes the RGB image and timestamp. Additionally, $y_k$ consists of the estimated probability of a hand-object interaction $p_o$ in this frame and the three-dimensional locations of the hands $\vec{r}_l, \vec{r}_r \in \mathbb{R}^3$ if they are within the field of view.

The problem formulation in \ref{problem} allows us to decompose the estimation process into two distinct prediction problems. First, the aim is to determine the temporal interval of hand-object interactions and, secondly, to estimate the 6D pose of the object carried within this interval.

The interval predictions boil down to finding the start and end times of the interactions within the environment. We call an observation $y_k$ a positive observation $y_k^+$ iff the probability of the 2D hand-object interaction detector for this observation meets a specified threshold $p_o >\tau_o$.

The starting time can be determined by utilizing the 3D prior knowledge encapsulated within the initial scene graph representation. An observation $y_k$ is classified as a point of contact and thus the beginning of an interaction interval if it meets two criteria: (i) $y_k$ is positive ($y_k^+$) and (ii) a distance lower than~$\tau_d$ between the respective hand and the closest 3D object~$O$. To account for incorrect hand-object predictions, a minimum number of $\theta_\text{reg}$ positive observations $y_k^+$ is required and all hand locations in the look-ahead horizon need to be further away from object $O$ than the current hand position.

The main principle to determine the end time of the interaction is again relying on the image-based hand-object predictions. However, because the object position is changing, we lack the 3D distance metric used for start point discovery. To compensate for this lack of knowledge, we compute the past and future hand velocity (denoted as $v_{\text{prior}}$, $v_\text{{post}}$) as the average Euclidean displacements between consecutive hand positions in our buffer $B$ and look-ahead $H$, respectively.
The concept builds upon the natural assumption that placing an item leads to a hand velocity reduction, whereas the retrieval of an object typically increases hand velocity afterwards. In contrast, actions such as carrying an object tend to exhibit a more constant velocity throughout.

Before, the decision rule would simply require a certain number $\theta_{\text{reg}}$ of positive observations
in the look-ahead horizon $H$. This formulation is further extended through the introduction of a second threshold $\theta_\text{high}$, applicable to cases with a notable discrepancy in hand velocity 
denoted as $\delta_\text{diff}$:
\begin{equation} \label{eq:velocity_extension}
    \sum_{k\in H} \mathbb{I}_{\{y_k\}}^+ \geq 
    \begin{cases} 
    \theta_{\text{high}} & \text{if } \lvert v_{\text{prior}} - v_{\text{post}} \rvert > \delta_\text{diff} \\
    \theta_{\text{reg}} & \text{else}
    \end{cases}
\end{equation}

\subsection{Pose Estimation and Tracking} \label{sec:pose_est}

In between the start and end point, we then track for each time step $k$ how the 6DoF pose $\textbf{T}_k = [\textbf{R}_k|\vec{t}_k] \in \mathds{SE}(3)$ of the interacted object changes. For each object, we compute a deterministic and meaningful initial pose $\textbf{T}_0$. 


\noindent{\textbf{Orientation Tracking}}:
To estimate the rotation, we track visual key points on the object. We initialize the key points by projecting 3D points of our 3D scene graph that belong to the tracked object $O$ onto the RGB image at the start time. We use the 3D points from the semi-dense Aria point cloud, gathered through nearest-neighbour matching with the object instance mask. Note that the semi-dense reconstruction is based on visual SLAM, and therefore these points are already known to be good key points.\enlargethispage{\baselineskip}

We then employ a point-tracking algorithm to predict key point positions and visibility for the images within the interaction in an online fashion. Since the key points are initialized by projecting the 3D model onto 2D, the tracking provides a set of visible key points for each frame, with established 2D-3D correspondences to the 3D object model.
Given these correspondences,
a RANSAC-based PnP algorithm robustly solves the rotation $\textbf{R}_{k, \text{obj}\rightarrow \text{cam}}$ from the object to the camera coordinate system at time step $k$.
Multiplying with the rotation
of the known camera pose $\textbf{R}_{k, \text{cam}\rightarrow \text{world}}$ as in equation \ref{eq:rotation} yields the orientation of our object in world coordinates.
\begin{equation} \label{eq:rotation}
    \textbf{R}_{k, \text{obj}\rightarrow \text{world}} = \textbf{R}_{k, \text{cam}\rightarrow \text{world}} \cdot \textbf{R}_{k, \text{obj}\rightarrow \text{cam}}
\end{equation}
\textbf{Translation Tracking}:
To estimate object translation, we utilize hand tracking. For the initial frame, we can use the 3D prior of the scene to obtain the relative distance vector between the hand and the tracked object. Given the initial hand position $\vec{r}_{0, \{l,r\}}$ and the centroid of the object $\vec{c}_o$ in the world coordinate frame, we calculate $\vec{\Delta}_{0, \text{world}}=\vec{c}_o-\vec{r}_{\{l,r\}}$. Because this offset is anchored at the centroid, it suffices to apply the inverse rotation to transform the offset into the object's local coordinate frame.
\begin{equation}
    \vec{\Delta}_{\text{obj}} = \textbf{R}_{0, \text{obj}\rightarrow \text{world}}^T \cdot \vec{\Delta}_{0, \text{world}}
\end{equation}
Given that the hand keeps grasping the object at the same point, this local offset $\vec{\Delta}_{\text{obj}}$ remains invariant throughout the interaction. We refer to this assumption as `hand anchor' in our ablation experiments.
We
leverage
this observation in equation~\ref{eq:translation} by transforming $\vec{\Delta}_{obj}$ back to $\vec{\Delta}_{k, \text{world}}$ for each frame $k$ given the estimated 3D hand locations and the rotations of the object.\enlargethispage{\baselineskip}
\begin{equation} \label{eq:translation}
    \vec{t}_k = \vec{r}_{k, \{l,r\}} - \textbf{R}_{k, \text{obj}\rightarrow \text{world}} \cdot \vec{\Delta}_{obj}
\end{equation}
Hence, the object pose $\textbf{T}_k$ can be reconstructed for each time step by combining $\textbf{R}_{k, \text{obj}\rightarrow \text{world}}$ and $\vec{t}_k$. The object pose is updated by applying the inverse of the pose from the previous step and subsequently the most recent pose prediction.
Moreover, as long as a hand is regarded as being in contact with an object, we start a separate point tracking and object pose estimation for this respective hand only.
This is achieved by storing the initial offset to the tracked object as well as projecting the 3D object points \emph{per} hand, even when both hands grab the same object. These concepts facilitate the tracking of multiple interactions with the same or different objects and the concurrent updating of two objects.

\begin{table*}[t]
\setlength{\tabcolsep}{3pt}
\renewcommand{\arraystretch}{1.3} 
\centering
\caption{\textsc{\textbf{Comparison of tracking performance}} across different methods for various objects with about 10 pick-and-place actions per object. The efficacy of each method is assessed based on three key metrics: RMSE of the object position (in centimetres) and orientation (in degrees) as well as ADD score (in percentage, 10\% object diameter threshold). 
The lowest position and orientation errors are highlighted in \textbf{bold \textcolor{first}{$\CIRCLE$}}, whereas the second-best values are \underline{underlined} \textcolor{second}{$\CIRCLE$}. Similarly for ADD, we bold the highest and underline the second-highest values. \emph{Mean} represents the average tracking error across all objects.}
\begin{tabular}{l|ccc|ccc|ccc|ccc|ccc}
\Xhline{2\arrayrulewidth}

\multirow{2}{*}{\parbox[c]{1.8cm}{\textbf{Objects}}} & \multicolumn{3}{c}{BundleTrack~\cite{bundletrack}} & \multicolumn{3}{c}{BundleSDF~\cite{bundlesdf}} & \multicolumn{3}{c}{FoundationPose~\cite{foundationpose}} & \multicolumn{3}{c}{Head Pose} & \multicolumn{3}{c}{Lost\&Found \emph{(ours)}} \\
\cline{2-16}

& \multicolumn{1}{c}{cm} & \multicolumn{1}{c}{$^\circ$} & \multicolumn{1}{c}{\%} & 
  \multicolumn{1}{c}{cm} & \multicolumn{1}{c}{$^\circ$} & \multicolumn{1}{c}{\%} & 
  \multicolumn{1}{c}{cm} & \multicolumn{1}{c}{$^\circ$} & \multicolumn{1}{c}{\%} & 
  \multicolumn{1}{c}{cm} & \multicolumn{1}{c}{$^\circ$} & \multicolumn{1}{c}{\%} & 
  \multicolumn{1}{c}{cm} & \multicolumn{1}{c}{$^\circ$} & \multicolumn{1}{c}{\%} \\

\hline

Carton  &  \nd{$7.77$}  &  $40.02$  &  \nd{$35.44$}  &  $9.24$  &  $29.92$  &  $34.62$  &  $9.07$  &  $35.28$  &  $26.10$  &  $7.84$  &  \nd{$18.80$}  &  $34.95$  &  $\fs{2.64}$  &  $\fs{5.13}$  &  $\fs{74.35}$   \\
Organizer  &  \nd{$4.87$}  &  $19.45$  &  \nd{$49.19$}  &  $11.28$  &  $\fs{6.25}$  &  $25.28$  &  $14.02$  &  $7.52$  &  $23.78$  &  $12.68$  &  $13.72$  &  $11.04$  &  $\fs{3.80}$  &  \nd{$6.53$}  &  $\fs{50.38}$   \\
Frame  &  $6.00$  &  $34.89$  &  $35.85$  &  $5.64$  &  $24.85$  &  $48.65$  &  $8.46$  &  $49.33$  &  $39.35$  &  \nd{$3.27$}  &  \nd{$12.04$}  &  \nd{$59.96$}  &  $\fs{1.99}$  &  $\fs{4.10}$  &  $\fs{90.39}$   \\
Clock  &  $6.43$  &  $52.33$  &  $10.18$  &  $6.88$  &  $64.37$  &  $6.34$  &  $8.86$  &  $67.96$  &  $1.82$  &  \nd{$3.76$}  &  \nd{$11.53$}  &  \nd{$21.89$}  &  $\fs{3.15}$  &  $\fs{6.07}$  &  $\fs{32.65}$   \\
Ball  &  \nd{$7.46$}  &  $36.93$  &  $4.38$  &  $15.77$  &  $21.69$  &  $4.02$  &  $13.97$  &  $29.85$  &  $3.34$  &  $15.03$  &  \nd{$15.66$}  &  \nd{$8.86$}  &  $\fs{4.60}$  &  $\fs{7.08}$  &  $\fs{13.29}$   \\
Plant  &  $5.45$  &  $29.66$  &  $38.77$  &  \nd{$4.77$}  &  $20.10$  &  $43.37$  &  $10.76$  &  $75.62$  &  $15.80$  &  $\fs{4.61}$  &  \nd{$11.55$}  &  \nd{$51.82$}  &  $4.87$  &  $\fs{5.24}$  &  $\fs{69.62}$   \\
Basket  &  \nd{$12.56$}  &  $27.46$  &  \nd{$41.35$}  &  $20.99$  &  \nd{$14.06$}  &  $28.30$  &  $22.83$  &  $26.97$  &  $22.21$  &  $18.80$  &  $24.16$  &  $17.24$  &  $\fs{5.26}$  &  $\fs{10.41}$  &  $\fs{75.14}$   \\
Water Can  &  $18.66$  &  \nd{$23.09$}  &  $43.74$  &  $21.60$  &  $23.88$  &  \nd{$45.62$}  &  $21.75$  &  $68.70$  &  $28.41$  &  $\fs{10.63}$  &  $28.90$  &  $26.84$  &  \nd{$15.58$}  &  $\fs{9.65}$  &  $\fs{69.91}$   \\
Shoe  &  \nd{$13.22$}  &  $23.95$  &  \nd{$20.62$}  &  $15.76$  &  $34.69$  &  $11.50$  &  $19.94$  &  $59.50$  &  $9.42$  &  $13.80$  &  \nd{$22.19$}  &  $11.18$  &  $\fs{12.28}$  &  $\fs{15.91}$  &  $\fs{30.07}$   \\
\hline
\textbf{Mean}  &  \nd{$9.16$}  &  $31.98$  &  \nd{$31.06$}  &  $12.44$  &  $26.65$  &  $27.52$  &  $14.41$  &  $46.75$  &  $18.91$  &  $10.05$  &  $\nd{17.62}$  &  $27.09$  &  $\fs{6.02}$  &  $\fs{7.79}$  &  $\fs{56.20}$   \\

\Xhline{2\arrayrulewidth}
\end{tabular}\vspace{-2mm}
\label{tab:transformed_experiments}
\end{table*}

\subsection{Implementation Details}
For scene graph creation, we capture an initial high-resolution scan of the environment using an iPad Pro and a 3D Scanner app~\cite{scanner}.
The scene is then segmented into its instances by Mask3D~\cite{mask3d}.

We capture human interactions with the head-mounted Aria glasses. For each recording, we utilize the 6DoF closed-loop slam trajectory of the Aria device, a semi-dense point cloud consisting of key points from its SLAM system, and the estimated 3D locations of wrist and palm tracking data. The integrated RGB camera captures 30 frames per second with a resolution of $1408\times1408$ pixels. Visual localization is carried out with a fixed fiducial marker~\cite{aruco} that was placed inside the scene. Alignment of the marker poses and further refinement via ICP~\cite{icp} allow us to achieve a common reference frame of our scene graph data structure and the Aria recording.

To detect hand-object interactions from 2D image observations, we use the pre-trained model by Shan et al.~\cite{Shan20} with a threshold $\tau_o~=~0.5$.
Both, the buffer $B$ and the look-ahead horizon $H$ in our sliding window, have a size of 8, inducing an overall delay of $1.15$s between input and output. 

For an observation to be classified as a starting point of a 3D hand-object interaction, the state itself has to be \emph{positive}, confirm with the distance threshold of $\tau_d=10$~cm and require at least $\theta_{\text{reg}}=4$ \emph{positive} observations in the look-ahead, or $\theta_{\text{high}}=6$ iff the absolute discrepancy of $v_{\text{prior}}$ and $v_{\text{post}}$ is greater than $\delta_\text{diff}=0.025~\frac{\text{m}}{\text{s}}$. The same rule applies, except for the distance threshold, to all other observations within the interaction interval.
For point tracking, we integrate an online version of CoTracker2~\cite{cotracker}. Its window size of $8$ allows for seamless integration with our employed look-ahead horizon, making the pose of the object readily available, once a particular frame is processed.

The aforementioned hyperparameters have been selected in training scenes that are not part of the experiments, and remain constant in this configuration throughout all of the following sections. 

\section{Experiments \& Results}
\subsection{Experimental Setup}
We conduct quantitative experiments to evaluate the tracking accuracy of our proposed method. The experiments are carried out within a furnished area that is designed to resemble a typical living room. The furniture includes items such as shelves and chairs, as well as various smaller movable objects. The full setup is depicted in Fig. \ref{fig:teaser_fig}.
\enlargethispage{\baselineskip}

To capture the three-dimensional trajectories of each displaced object, the environment is equipped with a Vicon motion capture system. Objects are tracked using mocap markers, which are rigidly attached to the objects’ surfaces. This specific point is then selected manually within the designated object frame of the scene graph, to ensure that the initial pose of the ground truth (mocap) and the predicted trajectory are as closely aligned as possible. Each recording consists of a single pick-and-place action that is performed by a person wearing Aria glasses. For each object, around 10 different trajectories are recorded, with each recording lasting between 10 and 20 seconds. Two instances of interaction tracking failure were observed in the 96 recordings. In both cases, the end of the interaction interval was mispredicted and the object was incorrectly identified as being dropped off mid-way through the sequence. All methods including baselines are equally affected by this.

It is important to note that the ground truth is recorded in a fixed frame, while the evaluated visual methods generate tracking estimates within an arbitrary Aria reference frame, which is set by the head pose at the beginning of each recording. To achieve both spatial and temporal calibration between these frames, we mount additional mocap markers onto the Aria glasses. Calibration is then performed by aligning the headset pose in the Aria reference frame with the corresponding trajectory tracked by the Vicon system. This alignment ensures that both systems are synchronized, providing a unified reference for accurate trajectory analysis.\enlargethispage{\baselineskip}


\subsection{Baselines}
As a first baseline, we rely solely on the head poses given by the Aria glasses for the object's orientation and the estimated 3D hand location for the object’s translation. This heuristic assumes that the carried object remains in the same orientation within the camera frame and omits any further pose processing. We refer to it as the Head Pose baseline, and it serves as an initial indicator of the attainable degree of accuracy.

Additionally, we compare multiple state-of-the-art methods for 6DoF pose estimation of unseen objects.
Our input for the following baselines initially consists of the egocentric camera recording within the identified tracking interval. This time interval remains constant for all baselines and our method.

All following methods require RGB-D data and object masks as input. To bridge this domain gap, we employ Metric3Dv2~\cite{metric3dv2} with known Aria camera intrinsics to generate metric depth estimates. Additionally, we utilize our 3D prior and reproject the 3D mesh of the object onto the first image of our tracking sequence. The object has not been moved yet, so this yields an accurate initial mask.

First, we implement variants of BundleSDF~\cite{bundlesdf} and BundleTrack~\cite{bundletrack} that also require masks for the rest of the interaction interval. To achieve this, we employ SAM2~\cite{sam2} and follow an iterative procedure, where the mask of the previous iteration is used as a prompt for the subsequent frame.
As a third baseline, we employ the most recent FoundationPose. An initial mask estimate is sufficient for this approach. However, we need to prompt this method with a 3D mesh of the object, which is obtained by selecting all mesh vertices of our 3D prior that belong to the tracked object instance.

Each baseline estimates the object pose $\textbf{T}_{k, \text{obj}\rightarrow \text{cam}}$ in the camera coordinate system. Given the head pose trajectory, we can then deduce the object pose $\textbf{T}_k$ in the world frame for each baseline. 
\subsection{Tracking Accuracy} \enlargethispage{\baselineskip}
\begin{figure*}[t]
    \centering
    \includegraphics[width=1.0\linewidth]{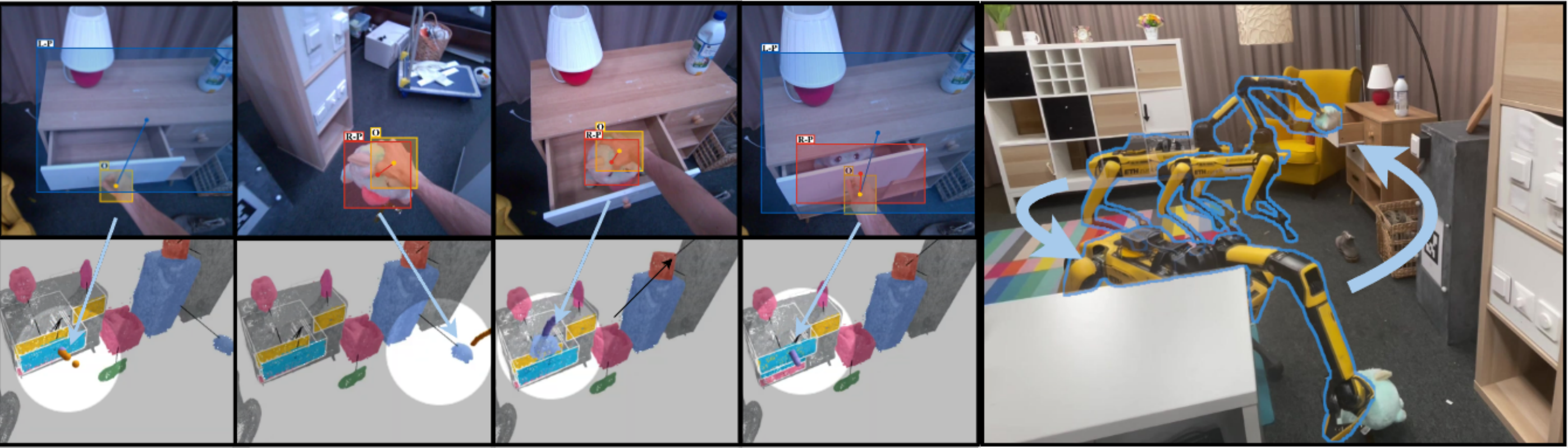}
    \caption{\textbf{Teach \& Repeat Experiment.} We showcase how Lost \& Found can help to record reoccurring motion primitives. In this example (from left to right), a human agent opens the top-left drawer of the small cabinet and grabs the blue toy from the other side of the room. The toy is then stored inside the drawer. To conclude the action, the drawer is closed again. We demonstrate that our method can be seamlessly integrated into robotic systems that are then capable of replaying the tracked interaction (on the right).
     \vspace{-5mm}
    }
    \label{fig:teach_repeat}
\end{figure*}

We first provide a detailed analysis of the tracking accuracy and robustness across different objects in Tab. \ref{tab:transformed_experiments}, always analyzing full trajectories. While full trajectory tracking might be relevant to e.g. imitation learning, our presented downstream applications are only dependent on the final object pose after interaction. We therefore provide final pose error and other overall performance metrics in Tab.~\ref{tab:summary_metrics}.

In Tab. \ref{tab:transformed_experiments} we report RMSE for the position (in centimetres) and orientation (in degrees) of each object trajectory, which includes items such as, among others, a carton, picture frame, and clock, to represent typical indoor objects with varying shapes and textures. On the recorded test set, we reduce the translation error by $34$\%  and the rotational error by $56$\% compared to the second best approach with a mean of $6.02$~cm and $7.79$~°, respectively.
Our approach notably improves orientation estimation, as reflected in the ADD~\cite{posecnn} score, which accounts for both translation and orientation errors by comparing all object points.

Notably, the Head Pose, which uses the wearer’s head orientation as a proxy of object rotation, shows competitive results, indicating a correlation between head and object movements. Only when the object is tilted within the camera frame, does the Head Pose heuristic suffer, hence the notable degradation of the orientation estimates compared to our method. Still, as this heuristic relies mostly on the very accurate camera poses, there are almost no extreme failure cases (e.g. PnP mismatch in our method), explaining good results throughout all scenes and even outperforming all other methods for the plant and watering can in terms of translation error.

The remaining methods show higher and more variable error rates,  exhibiting particularly high orientation errors, especially on objects with complex geometries, such as the clock and plant. These tracking approaches appear to be negatively affected by the lack of depth information, which modern monocular depth estimators cannot sufficiently replace yet.
They excel and perform state-of-the-art in lab settings, but the noisy depth estimates paired with the egocentric viewpoint (clutter and sudden viewpoint changes) degrade the performance a lot.
\begin{table}[t]
\setlength{\tabcolsep}{3pt}
\renewcommand{\arraystretch}{1.2} 
\centering
\caption{\textbf{\textsc{Overall Metrics}:} Performance metrics across all objects in the test set. We state ADD-S (\%) for 10\% of the object diameter, percentage of poses in the whole trajectory within a $5$cm and $5^\circ$ error threshold as well as translational (cm) and rotational ($^\circ$) error of the final end pose and the inference frames per second\vspace*{-5mm}}. 
\resizebox{\columnwidth}{!}{
\begin{tabular}{l c c c c c}
\toprule
{Metric} & \multicolumn{1}{c}{ADD-S ($\uparrow$)} & \multicolumn{1}{c}{Acc$_{5cm,5^\circ}$ ($\uparrow$)}  & \multicolumn{1}{c}{$\textbf{T}_{\text{end}}$ ($\downarrow$)} & \multicolumn{1}{c}{$\textbf{R}_{\text{end}}$ ($\downarrow$)} & \multicolumn{1}{c}{FPS ($\uparrow$)}\\
\midrule
BundleTrack & \nd{$73.81$} & \nd{$16.21$} & \nd{$12.01$} & $42.79$ & (1.72)\\
FoundationPose & $44.56$ & $10.34$ & $17.97$ & $65.00$ & $2.56$\\
BundleSDF & $60.69$ & $14.18$ & $15.60$ & $34.56$ & $1.45$\\
Head Pose & $68.26$ & $13.51$ & $12.33$ & \nd{$24.42$} &  $\fs{7.48}$\\
Lost\&Found \emph{(ours)} & $\fs{88.10}$  & $\fs{53.05}$  & $\fs{8.46}$ & $\fs{10.91}$ & $\nd{6.95}$\\
\bottomrule
\end{tabular}
}\vspace{-3mm}
\label{tab:summary_metrics}
\end{table}

In the case of BundleSDF, pose estimation is predominantly reliant on finding a consistent neural 3D reconstruction of the object. Inaccurate depth estimates result in erroneous geometrical cues in this regard, which subsequently influence the resulting pose predictions. FoundationPose relies on the object mesh, which is often incomplete and scattered in our real-world setup. In combination with the misaligned depth, this could explain the difficulties of matching the correct object pose. BundleTrack achieves the second-best results among the evaluated approaches. It relies more on robust feature matching of key points for the pose estimation and hence suffers less from incorrect depth estimates.
In Lost \& Found, we also make use of this idea. However, instead of a classical key point detection and feature matching approach, we employ a readily available point tracking algorithm. Additionally, we make use of the 3D prior given as our scene graph representation. With this, we know the ground truth spatial dependency of the 2D feature points, giving us a direct way to integrate the 3D geometry in the whole tracking process. The introduced temporal coherence of the feature points paired with the injected 3D knowledge has proven to give more accurate and reliable pose predictions. These adaptions are not only reflected in the quantitative study but also lead to visibly smoother 6DoF trajectories with less strong flickering artifacts compared to the baselines. 
\enlargethispage{\baselineskip}

Additional quantitative results for the test set are presented in Table~\ref{tab:summary_metrics}, where we summarize the average performance across all object categories, along with the average processing speed measured in frames per second. Specifically, the table includes the average ADD-S and Acc$_{5cm,5^\circ}$ scores, which are evaluated on the \emph{entire} trajectory. 
Furthermore, we report the pose error at the end of the trajectory, which is of great importance for our downstream robotic pick-and-place applications (Sec. \ref{sec:robo_examples}).
We measure inference time of the methods themselves, without times for state estimation of camera pose and hand pose that all methods rely on, e.g. to detect the interaction interval. Such state estimation is usually built-in with the headset firmware and therefore reasonable to assume available.
All inference times are measured on an Nvidia GeForce RTX 4090 with the exception of BundleTrack which only ran on an older 3090 GPU. Lost~\&~Found demonstrates an efficient processing rate on the test set, achieving an average throughput of $6.95$~fps - only slightly slower than the Head Pose baseline. 
In contrast, BundleTrack, BundleSDF, and FoundationPose all have lower frame rates. 
Notably, for all these methods, the primary computational overhead stems from input processing, specifically, inferring hand-object interactions, generating masks and predicting monocular depth. 


\vspace{-3pt}
\subsection{Ablation} 
\label{sec:experiments_ablation} All methods except the Head Pose baseline yield the complete 6DoF pose $\textbf{T}_{k, \text{obj}\rightarrow \text{cam}}$ in the camera frame. Hence, we could directly obtain the pose of the object $\textbf{T}_k$ without using the estimated 3D hand location as an anchor. A respective comparison of each method's accuracy only using the predicted pose (\xmark) or utilizing the 3D hand location (\cmark) as an additional input signal is depicted in Tab. \ref{tab:ablation_study}. \enlargethispage{\baselineskip}

The results reaffirm the limitations of the baselines in accurately predicting the pose. Here, omitting the hand location has a clearly negative effect. In contrast, our method achieves competitive results even without this anchor, allowing us to track object poses even when the grasping positions of the hands change. In terms of translation error, this approach even outperforms our full method by a slight margin, thereby demonstrating the suitability of our approach also in the absence of 3D hand locations.
\subsection{Robotic Downstream Tasks} \label{sec:robo_examples}
This section demonstrates the importance of a transformable scene representation for robot navigation tasks.
In this setup, we combine the tracking of objects in the scene with the detected drawers. We use our same pipeline to track the drawer pose, but restrict their movement to the axis defined by the surface normal of the drawer's front face.
Given our initial state, we can perform arbitrary interactions within the scene, captured solely by Aria glasses, and then perform informative robotic manoeuvres in the same environment.

\noindent{\textbf{Object Retrieval:}} We select a particularly challenging scenario in which an object is stored in one of the drawers during the interaction. The occlusion of the object precludes the naive solution of a rescan. However, our scene graph structure contains the updated state of the scene with the object connected through a 'contains' edge with the correct drawer.
This allows a mobile manipulator to successfully find and retrieve the object, despite its lack of visibility at the moment of the query.\enlargethispage{\baselineskip}
\begin{table}[t]
\setlength{\tabcolsep}{3pt}
\renewcommand{\arraystretch}{1.0}
\centering
\caption{\textbf{\textsc{Ablation Study}:} Each method is evaluated with the estimated pose as the only proxy (\xmark) or with the 3D hand location as additional input (\cmark). Because the orientation remains invariant for the two configurations, we report $\textbf{T}_\text{traj}$ of the whole trajectory (in cm), ADD and ADD-S score (both in percent and for 10\% diameter threshold).}
\begin{tabular}{l c c c c}
\toprule
Method & Hand Anchor & $\textbf{T}_\text{err}$ ($\downarrow$) & ADD ($\uparrow$) & ADD-S ($\uparrow$) \\
\midrule
Head Pose & \cmark & $10.07$ & $26.91$ & $68.26$ \\
\midrule
\multirow{2}{*}{BundleTrack} & \xmark & $19.31$ & $9.61$ & $35.22$ \\
& \cmark & $9.05$ & $30.99$ & $73.81$ \\
\midrule
\multirow{2}{*}{FoundationPose} & \xmark & $41.73$ & $6.04$ & $24.49$ \\
& \cmark & $14.33$ & $18.70$ & $44.56$ \\
\midrule
\multirow{2}{*}{BundleSDF} & \xmark & $19.17$ & $10.65$ & $34.93$ \\
& \cmark & $12.38$ & $27.28$ & $60.69$ \\
\midrule
\multirow{2}{*}{Lost\&Found \emph{(ours)}} & \xmark & $\fs{5.87}$ & $\nd{48.97}$ & $\nd{82.62}$ \\
& \cmark & $\nd{5.90}$ & $\fs{55.97}$ & $\fs{88.10}$ \\
\bottomrule
\end{tabular}\vspace{-5mm}
\label{tab:ablation_study}
\end{table}

\noindent{\textbf{Teach \& Repeat:}}
Similarly, we can use the tracking sequence to replay interactions in a real-world setting. By providing the start and end pose of the particular object trajectory to the robot, it is capable of computing object grasps and planning body movements that align with the carried-out task. We not only showcase this for simple pick-and-place actions but also for more complex scenarios when objects are stored in or taken out of drawers (Fig. \ref{fig:teach_repeat}). Such a teach \& repeat capability could prove useful in defining reoccurring motion primitives, also for people with a less technical background.

\section{Limitations \& Future Work}
As with other pose trackers, Lost \& Found is inherently limited to 6DoF pose estimation of \emph{rigid} objects. However, this rigid-body assumption is only applicable to a subset of objects with which we interact in the real world. While our robotic demonstrations include the case of the plush-cow, for instance, that we can still track until it is dropped in the drawer, our method would likely struggle under more severe deformations.
Moreover, the object must remain in the perceptual field of the egocentric observations at all times to make our approach effective (110\degree~for the Aria RGB camera). Although this is straightforward to impose, it does not align with how we typically transport certain objects. This limitation may be lifted as point trackers improve in recovering from occlusions and tracking loss.\enlargethispage{\baselineskip}

Naturally, limitations within the underlying methods do propagate to our method. For instance, solving the depth ambiguity that 2D observations yield would result in a performance leap for detecting hand-object interactions - a crucial building block in the Lost \& Found pipeline that is still prone to error.

Our experiments only focus on changes that occur within the egocentric observations. Incorporating updates through eventual re-scans or adding physics-based constraints such as 'objects should always be situated on the floor or on another entity'  could correct wrong tracking results and ensure the long-term consistency of the scene graph.

In addition to improving the fundamental techniques, it could be advantageous to model the uncertainty associated with the tracked object. Such an approach could handle ambiguous observations that can be resolved later with higher certainty.
While we demonstrated possible applications of dynamic scene graphs with robotic agents in \ref{sec:robo_examples}, a comprehensive framework that effectively integrates both domains could provide valuable resources to the research community.

\section{Conclusion}
In this work, we proposed Lost \& Found, an approach to accurately detect and track object interactions in 3D environments. The framework integrates a 6DoF pose tracking method that handles egocentric viewpoints and limited domain knowledge (missing depth) more robustly than state-of-the-art. We employ this object interaction tracking on a transformable scene graph data structure and showcase how robotic systems benefit from the acquired dynamic knowledge.

\section*{Acknowledgment}
We thank Tim Engelbracht for assisting with the robot.
\bibliographystyle{IEEEtran}
\bibliography{IEEEabrv,references}
\end{document}